\documentclass[12pt]{spieman}  
\usepackage{amsmath,amsfonts,amssymb}
\usepackage{graphicx}
\usepackage{setspace}
\usepackage{tocloft}
\usepackage{multirow}

\title{Pore detection in high-resolution fingerprint images using Deep Residual Network}

\author[a]{Vijay Anand}
\author[a*]{Vivek kanhangad}
\affil[a]{Discipline of Electrical Engineering, Indian Institute of Technology Indore, Indore, India, 453552}

\cftpagenumbersoff{figure}
\cftpagenumbersoff{table} 
\begin{document} 
\maketitle

\begin{abstract}
This letter presents a residual learning-based convolutional neural network, referred to as DeepResPore, for detection of pores in high-resolution fingerprint images. Specifically, the proposed DeepResPore model generates a pore intensity map from the input fingerprint image. Subsequently, the local maxima filter is operated on the pore intensity map to identify the pore coordinates. The results of our experiments  indicate that the proposed approach is effective in extracting pores with a true detection rate of $94.49$\% on Test set I and $93.78$\% on Test set II of the publicly available PolyU HRF dataset. Most importantly, the proposed approach achieves state-of-the-art performance on both test sets.  
\end{abstract}

\keywords{pore detection, high-resolution fingerprint, deep residual network}

{\noindent \footnotesize\textbf{*}Vivek Kanhangad,  \linkable{kvivek@iiti.ac.in} }

\begin{spacing}{2}   

\section{Introduction}
\label{sect:intro}  
Pores, which constitute the extended fingerprint features, are generally observable in high-resolution fingerprint images \cite{jain2007pores}. The effectiveness of pore features for high-resolution fingerprint based biometric recognition has been explored and the features have been found to carry sufficient discriminating power.
They have also been shown to be effective for recognition using partial fingerprint images, which may not contain sufficient level-2 features \cite{Zhao20102833}. 
Studies \cite{jain2007pores,Zhao20102833} have also shown that the fusion of pore and level-2 features  leads to enhanced recognition performance. However, it is imperative that the pore coordinates are detected accurately.
The existing pore detection techniques can be broadly classified into feature-based techniques \cite{jain2007pores,Zhao20102833,Teixeira,Lemes} and learning-based techniques \cite{LABATI2017,Deep_pore}. Since deep convolutional neural networks (CNN) have achieved state-of-the-art performance for various computer vision tasks, researchers have focused on designing learning-based techniques for  detecting pores in high-resolution fingerprint images. Table \ref{all_cnn} presents a summary of such techniques.
Labati \emph{et al.} \cite{LABATI2017}  developed an approach for pore detection that employs a shallow CNN to generate a pore intensity map, which is thresholded to obtain the pore coordinates. Their approach \cite{LABATI2017}, however, does not provide any improvement over the existing feature-based pore detection techniques. Recently, Jang \emph{et al.} \cite{Deep_pore} presented a pore detection technique using a deep CNN that consists of 10 learnable layers. The pore intensity map generated by their network is processed to detect the pore coordinates. Their  approach \cite{Deep_pore} provides a considerable improvement in pore detection accuracy over the existing approaches.
However, DeepPore network presented in \cite{Deep_pore} has a plain CNN architecture  with only 10 learnable layers. 
Furthermore, their  approach has been evaluated on a very small test set containing only 6 fingerprint images. The contribution of this work is a residual learning \cite{Resnet} based CNN, referred to as DeepResPore, for  detecting pores in high-resolution fingerprint images. Specifically, we have designed an  18-layer  network containing 8 residual blocks inspired by ResNets \cite{Resnet}.
The proposed network has been trained on a large labeled dataset containing 210,330 patches and evaluated on multiple test sets to ascertain its performance. 
\begin{table}[ht!]
\centering
\caption{Learning-based approaches for automatic pore detection in \protect \\high-resolution fingerprint images}
\label{all_cnn}
\begin{tabular}{|c|c|c|c|c|}
\hline
\tiny \textbf {Study}    & \begin{tabular}[c]{@{}c@{}}\tiny \\ \tiny \textbf{Number of } \\ \tiny \textbf{CNN} \\ \tiny \textbf {layers}\end{tabular} & \begin{tabular}[c]{@{}c@{}}\tiny \textbf{Training} \\ \tiny \textbf{data}\end{tabular} & \tiny \textbf{Test data} & \tiny \textbf{Remarks} \\ \hline
\begin{tabular}[c]{@{}c@{}}\tiny {Labati  \emph{et al.}} \cite{LABATI2017}\end{tabular}   & \begin{tabular}[c]{@{}c@{}}\tiny {$2$ layers} \end{tabular}        &\begin{tabular}[c]{@{}c@{}}\tiny $24$ fingerprints \end{tabular}            & \begin{tabular}[c]{@{}c@{}}\tiny $6$ fingerprints  \end{tabular}        &\begin{tabular}[c]{@{}c@{}}\tiny {Plain and shallow  network};  \\ \tiny{without non-linear activation}; \\ \tiny{small train and test set}\end{tabular}       \\ \hline
\tiny DeepPore \cite{Deep_pore}& \begin{tabular}[c]{@{}c@{}}\tiny $10$ layers \end{tabular}            & \begin{tabular}[c]{@{}c@{}}\tiny $89,250$ patches   \end{tabular}           &\begin{tabular}[c]{@{}c@{}}\tiny $6$ fingerprints  \end{tabular}        & \begin{tabular}[c]{@{}c@{}}\tiny{Plain network};  \\   \tiny{small test set}\end{tabular}        \\ \hline
\tiny Proposed & \begin{tabular}[c]{@{}c@{}}\tiny $18$ layers \end{tabular}          & \begin{tabular}[c]{@{}c@{}}\tiny $210,330$ patches    \end{tabular}            & \begin{tabular}[c]{@{}c@{}}\tiny Test Set I, \tiny Test set II \\ \tiny( $30$ fingerprints each)  \end{tabular}         & \begin{tabular}[c]{@{}c@{}}\tiny{DAG and deep  network};  \\ \tiny{with residual learning}; \\ \tiny{large train and test set}\end{tabular}        \\ \hline
\end{tabular}
\vspace*{-5mm}
\end{table}

\section{Proposed Methodology}

In this letter, we present an   automated   pore   detection   methodology   that  employs  a  customized residual learning-based CNN model referred to as DeepResPore. The proposed methodology involves the following two major steps. Firstly,  a pore intensity map is generated from  the input fingerprint image by employing  DeepResPore. Secondly, the pore intensity map is processed to obtain estimates of the pore coordinates. 
We have developed a residual learning-based CNN model for the following reasons. Firstly, the residual learning provides a large receptive field, which is expected to result in increased representational power suitable for pixel-level predictions. Secondly, the residual learning leads to faster convergence and yields less training error compared to the plain stacked convolutional layers \cite{Resnet}. 
Table \ref{CNN} presents the detailed architecture of DeepResPore. As can be observed, the proposed network contains 18 layers with 8 residual blocks. Thus, there are a total of 8 shortcut connections in DeepResPore. Specifically, we have employed $1\times1$ convolutional and the identity shortcut connections in an alternating manner. The input to DeepResPore is a fingerprint patch of size $80\times80$ pixels and the output is the corresponding pore intensity map of the same size. In order to obtain a feature map of the same size as that of the input,  we have not performed pooling operation in the proposed  network. Pooling operation could also have adversely affected the pore coordinate estimation.
At the end of network, we have introduced  a convolutional layer  containing a single filter to provide a pore intensity map. All convolutional layers in DeepResPore  perform convolution with a stride of one and  zero padding is employed at every stage to maintain the size of the feature map. 
Furthermore, all convolutional layers, except the last one, perform convolution  followed by batch normalization  and ReLU activation. 
Finally, the last convolutional layer of DeepResPore provides the pore intensity map.
During the training phase, the network is trained end-to-end in a supervised learning manner with the objective of minimizing the value of the loss function. The output of DeepResPore  is a pore intensity map, in which pore coordinates are highlighted. Let $x$ be the input fingerprint and $y$ be the corresponding labeled pore intensity, where $x$ and $y$ are of the same size. Our network trains a model $f$ on the training set \{$x_{i}$,$y_{i}$\}, $i=1,2,\ldots,N $ to predict $\hat{y}=f(x)$, where $\hat{y}$ is the predicted pore intensity map and $N$ is the mini-batch size. At the regression output layer, the predicted pore intensity map is compared with the labeled pore intensity map using the following loss function:
\begin{equation}\label{loss}
  L_{CNN}=\frac{1}{N}\sum_{i=1}^{N}(y_{i}-\hat{y_{i}})^{2}
\end{equation}

\begin{table}[ht!]
\centering
\caption{Architecture of DeepResPore}
\label{CNN}
\begin{tabular}{|c|c|c|}
\hline
Layer name & output size & kernel                                                                        \\ \hline
conv1      & $80\times80$       & $7\times7$, $64$, stride 1, padding same                                                            \\ \hline
conv2\_x   &$80\times80$      & \Big[\begin{tabular}[c]{@{}c@{}}$3\times3$, $64$ \\ $3\times3$, $64$\end{tabular}\Big] $\times2$                  \\ \hline
conv3\_x   & $80\times80$      & \Big[\begin{tabular}[c]{@{}c@{}}$3\times3$, $128$ \\ $3\times3$, $128$\end{tabular}\Big] $\times2$                    \\ \hline
conv4\_x   &$80\times80$      & \Big[\begin{tabular}[c]{@{}c@{}}$3\times3$, $256$ \\ $3\times3$, $256$\end{tabular}\Big]  $\times2$                            \\ \hline
conv5\_x   &$80\times80$      & \Big[\begin{tabular}[c]{@{}c@{}}$3\times3$, $512$\\  $3\times3$, $512$\end{tabular}\Big]   $\times2$                           \\ \hline
conv6      & $80\times80$       & $3\times3$, $1$                                                                            \\ \hline
output     & $80\times80$       & Regression                                                                                \\ \hline
\end{tabular}
\vspace*{-3mm}
\end{table}

\section{Experimental results and discussion}
Since training a deep CNN from scratch requires a large labeled dataset, we have performed the experiments on  an annotated pore dataset that contains 138,643 annotated pores from 120  fingerprint images of resolution 1200 dpi and size $640\times480$ pixels \cite{tex_PAMI}. Out of  120  fingerprint images, we have used 90 fingerprint images from the first session to create the training dataset and remaining 30 fingerprint images from the second session to create the first test set referred to as Test set I (TS I). In  addition, we have evaluated the proposed model on a second test set referred to as Test set II (TS II) comprising the pore ground truth images from the PolyU high-resolution fingerprint database \cite{POLYU}.  TS II contains a total of 1,267 annotated pores from  30 fingerprint images of resolution 1200 dpi and size $320\times240$ pixels.
 To train the proposed DeepResPore, we have divided fingerprint images in the training set into overlapping patches of size $80\times80$ pixels with a step size of 10 pixels. Thus, we have a total of 210,330 fingerprint patches for training. We have then generated labeled pore intensity maps from the annotated pores. In this process, the pixels corresponding to the ground truth pore coordinates are marked as 1 in the labeled intensity map. Also, the pixels inside a 5-pixel radius from each of the ground truth pore coordinates are marked with labels between 0 and 1 and the remaining pixels are marked as 0, in the following manner \cite{Deep_pore}:
%

\begin{equation}
  L(i,j)=\begin{cases}
    1-\frac{d_{p}(i,j)}{5}, & \text{if $d_{p}(i,j)<5$}.\\
    0, & \text{otherwise}.
  \end{cases}
\end{equation}

where $L(i,j)$ is the label assigned to the pixel at $(i,j)$ and $d_{p}(i,j)$ is the distance between the pixel at $(i,j)$ and the ground truth pore coordinates.

\begin{figure}[t]
\centering
\includegraphics[width=.2\textwidth]{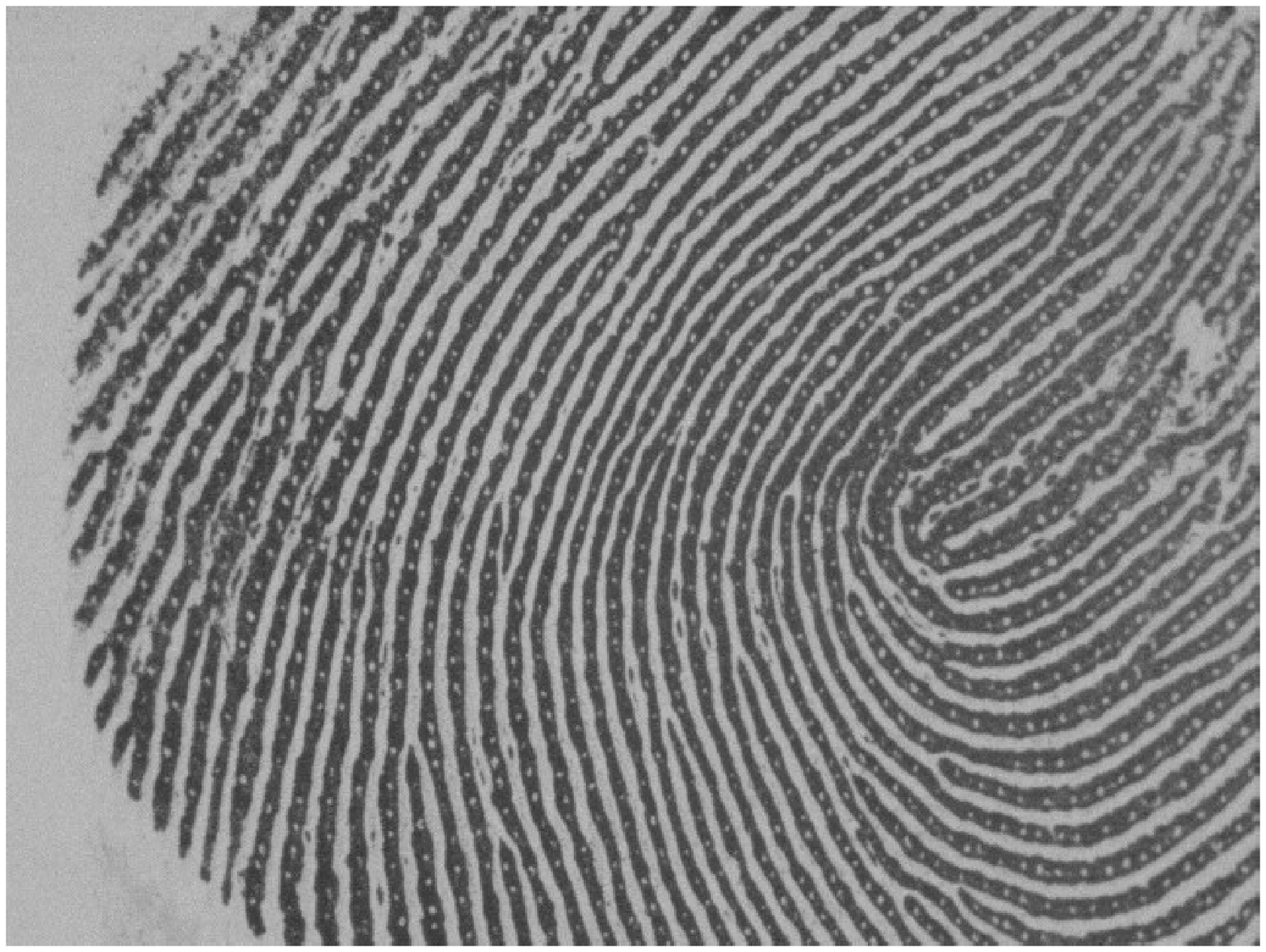}\quad \quad
\includegraphics[width=.2\textwidth]{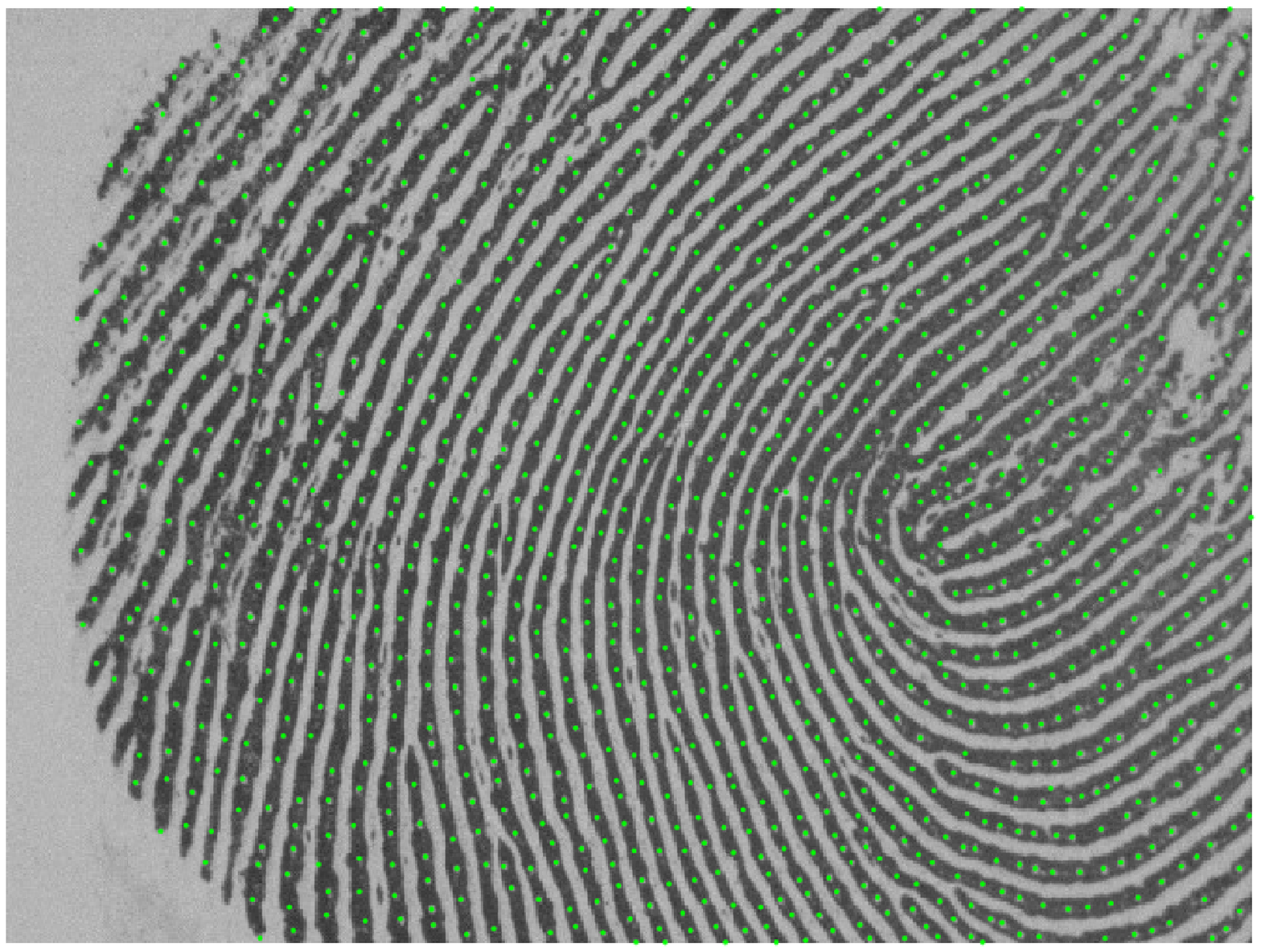}

\medskip

\includegraphics[width=.23\textwidth]{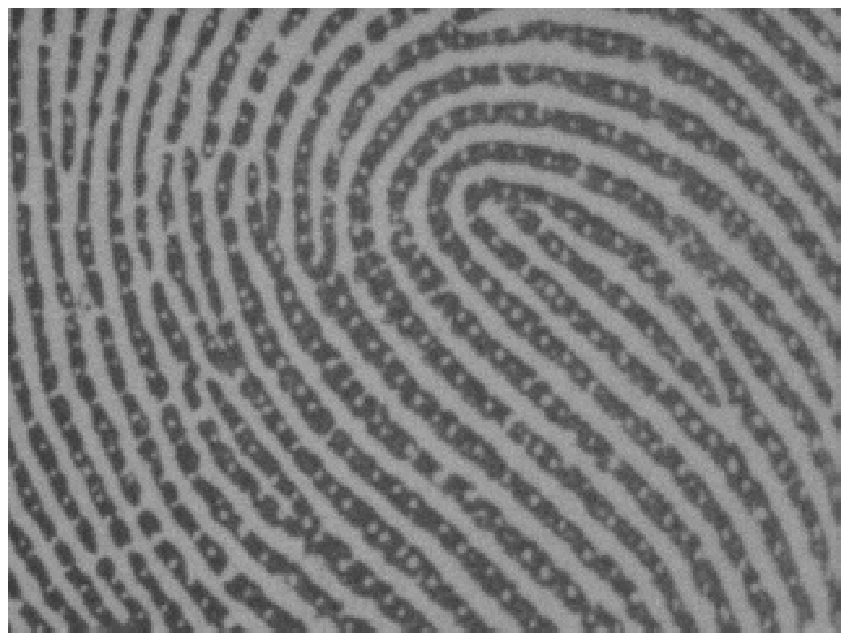}\quad
\includegraphics[width=.23\textwidth]{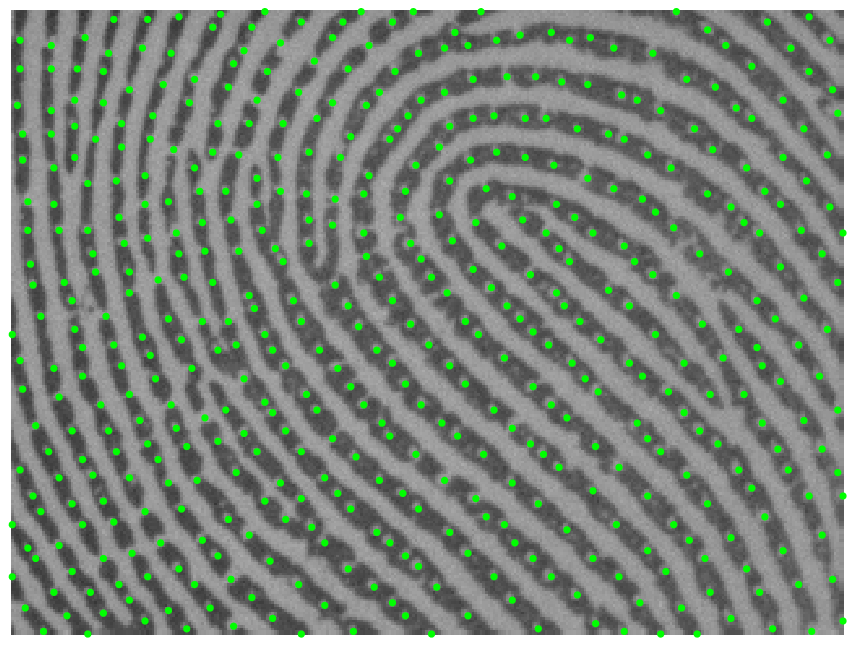}
\vspace*{-6mm}
\caption{Sample fingerprint images (left column) and the detected pores marked by green circles (right column)}
\label{finger_sample}

\end{figure}

In our experiments, the proposed DeepResPore has been trained in an end-to-end manner. The weights of the convolution layers have been initialized using a Gaussian function with zero  mean and  0.001 standard deviation. DeepResPore has been trained for 25 epochs (420,650 iterations) using a mini-batch size of 10  and a base learning rate of 0.001. We have employed the adaptive moment estimation (ADAM) solver \cite{ADAM} for optimizing the loss function, as it provides  adaptive learning rates.
During the testing phase, a test fingerprint image is first divided into non-overlapping patches of size $80\times80$ pixels, which are then fed to DeepResNet to obtain the corresponding pore intensity maps. All of these pore intensity maps are then combined to form a complete pore intensity map of the same size as the input fingerprint image. The generated pore intensity map highlights the pore coordinates and suppresses all other details in a fingerprint image. Therefore, the pore intensity map is processed further to obtain a binary pore map indicating the pore coordinates. 
Since pore pixels appear as local maxima in the pore intensity map, we have employed a spatial filtering based approach to estimate pore coordinates. Firstly, we have initialized two  2-D arrays representing the maximum value map and the binary pore map to zero.
We have then filtered the pore intensity map using maximum function with a window of size $5\times5$, which replaces every pixel in the maximum value map with the maximum of the pixel values in the neighborhood of the corresponding pixel in the pore intensity map. For every pixel in the pore intensity map, if its value is equal to the corresponding pixel value in the maximum  value map and greater than a predefined threshold $(th)$, then the corresponding pixel in the pore map is set to 1. On completion of this step, we obtain a binary pore map indicating the locations of the detected pores. Sample fingerprint images from each of the test sets and the  pores detected in them are shown in Fig. \ref{finger_sample}.

  To evaluate the performance of the proposed pore detection approach, two metrics namely, true detection rate $(R_{T})$ and false detection rate $(R_{F})$ are employed.  $R_{T}$ is defined as the ratio of the true detected pores to the pores present in the ground truth, while $R_{F}$ is defined as the ratio of the false detected pores to the total number of detected pores \cite{LABATI2017,Deep_pore}. In our experiments, a detected pore has been considered as a true pore if the Euclidean distance between the detected pore coordinates and the true pore (ground truth) coordinates is less than $RW/2$, where $ RW$ is the average width of the fingerprint ridges in the test set.
  For a comparative evaluation, we have also evaluated the performance of the existing CNN-based approaches \cite{LABATI2017,Deep_pore}. To this end, we have trained the three CNN models on the same dataset (\emph{i.e.} 210,330 fingerprint patches) by partitioning it into training and validation sets containing 168,264 and 42,066 image patches, respectively. We have then tested these models on two different test sets (TS I and TS II), each containing 30 fingerprint images. The results of our experiments are presented in Table \ref{performance}. To make a fair comparison, we have fixed the value of $R_{F}$ to 8.5 and computed the corresponding $R_{T}$ for the proposed and DeepPore based approach \cite{Deep_pore}.
 We have not been able to set $R_{F}$ to 8.5 for the approach presented in \cite{LABATI2017}, as it detects only a limited number of pores leading to a small value for $R_{F}$. 
As can be seen in Table \ref{performance}, the proposed DeepResPore-based approach provides a considerable improvement of  3.17\%  and 4.24\% points in $R_{T}$ over the current state-of-the-art approach \cite{Deep_pore} on Test set I and  II, respectively. In our experiments, the parameter $th$ is empirically set to 0.4 and 0.12  for Test set I and  II, respectively. 
All our experiments have been performed in MATLAB environment on a computer with 3.60 GHz Intel core i7-6850K processor, 48 GB RAM and Nvidia GTX 1080 8 GB GPU. Table \ref{compoutation} presents the average computation time required to generate a pore map for fingerprint images in  Test sets I and II.  As can be observed, the average time taken by the proposed approach to generate a pore map is marginally higher  than the current state-of-the-art approach \cite{Deep_pore}. 
In conclusion, our experimental results indicate that the proposed DeepResPore-based approach provides a considerable improvement in the pore detection accuracy with marginal increase in the computation time over the current state-of-the-art approach \cite{Deep_pore}.

\begin{table}[tbp]
\centering
\caption{Performance comparison with the existing methods}
\label{performance}
\begin{tabular}{|c|c|c|c|c|c|c|}
\hline
\multirow{2}{*}{\begin{tabular}[c]{@{}l@{}}Metric\end{tabular}} & \multicolumn{2}{l|}{~ ~ DeepPore \cite{Deep_pore}}               & \multicolumn{2}{l|}{~~Labati \emph{et al.}\cite{LABATI2017}}& \multicolumn{2}{l|}{~~DeepResPore }     \\
\cline{2-2}\cline{3-7}
                                                                              & {TS I }                      &  {TS II }& {TS I}    & {TSII}&  {TS I}& {TS II}  \\ \hline
~~$R_{T}$                                                                      & 91.32                            & 89.54       & 50.87                                                      & 52.21& \textbf{94.49}& \textbf{93.78}                    \\ \hline
~~$R_{F}$                                                                      & 8.5                             & 8.5      & 3.60                                                          & 1.62& 8.5& 8.5                 \\
\hline
\end{tabular}
\vspace*{-3mm}
\end{table}

\begin{table}[ht!]
\centering
\caption{Computational performance}
\label{compoutation}
\begin{tabular}{|c|c|c|c|c|c|c|}
\hline
\multirow{2}{*}{\begin{tabular}[c]{@{}l@{}}~~Time\\ (seconds)\end{tabular}} & \multicolumn{2}{l|}{~ ~ DeepPore \cite{Deep_pore}}               & \multicolumn{2}{l|}{~~Labati \emph{et al.}\cite{LABATI2017}}& \multicolumn{2}{l|}{~~DeepResPore }     \\
\cline{2-2}\cline{3-7}
                                                                              & {TS I }                      &  {TS II }& {TS I}    & {TSII}&  {TS I}& {TS II}  \\ \hline
~~$T_{avg}$                                                                      & 1.70                             & 0.61       & 0.30                                                     & 0.03 & {2.73}& {0.83}                    \\ \hline
\end{tabular}
\vspace*{-3mm}
\end{table}

\section{Conclusion}
In this letter, we have presented a methodology for detection of pores in high-resolution fingerprint images. Specifically, we have
developed a residual learning-based CNN named DeepResPore that generates  a pore intensity map for the input fingerprint image. 
Our experimental results suggest that the proposed DeepResPore  is  very effective in detecting pores in high-resolution fingerprint images.
Most importantly, the proposed DeepResPore model provides an improvement of  $3.17\%$ and $4.24\%$ points in true detection rates over the current state-of-the-art approach  on Test set I and II, respectively. 

\acknowledgments 
We thank Dr. Raoni F. S. Teixeira for providing us with the pore ground truth data for 120 fingerprint images. 


\bibliography{vijay_iet_letters}   
\bibliographystyle{spiejour}   





\end{spacing}
\end{document}